\pdfoutput=1

\documentclass[11pt]{article}
\usepackage[table]{xcolor}
\usepackage[final]{acl}

\usepackage{times}
\usepackage{soul}
\usepackage{url}
\usepackage[utf8]{inputenc}
\usepackage{caption}
\usepackage{graphicx}
\usepackage{amsmath}
\usepackage{amsthm}
\usepackage{booktabs}
\usepackage{algorithm}
\usepackage{algorithmic}
\usepackage[switch]{lineno}
\usepackage{natbib}
\usepackage {multirow}
\usepackage{graphicx}
\usepackage{amsfonts}
\usepackage{dsfont}
\usepackage{amsmath}
\usepackage{makecell}
\usepackage{bm}
\usepackage{arydshln}

\usepackage[T1]{fontenc}

\usepackage[utf8]{inputenc}

\usepackage{microtype}

\usepackage{inconsolata}

\title{Efficient $k$-Nearest-Neighbor Machine Translation\\ with Dynamic Retrieval}

\author{Yan Gao$^{1,2}$\thanks{These authors contributed equally.}, Zhiwei Cao$^{1,2}$\footnotemark[1], Zhongjian Miao$^{1,2}$, Baosong Yang$^3$, Shiyu Liu$^{2}$, Min Zhang$^4$, Jinsong Su$^{1,2}$\thanks{Corresponding author.}
\\
        $^1$School of Informatics,  Xiamen University, China \\ $^2$Key Laboratory of Digital Protection and Intelligent Processing of Intangible Cultural Heritage of \\ Fujian and Taiwan, Ministry of Culture and Tourism, China \\ $^3$Alibaba Group, China \\ $^4$Institute of Computer Science and Technology, Soochow University, China \\     \texttt{\small gaoyan@stu.xmu.edu.cn
    }~~
    \texttt{\small lines1@stu.xmu.edu.cn
    }~~
    \texttt{\small jssu@xmu.edu.cn
    }}

\begin{document}
\maketitle
\begin{abstract}
To achieve non-parametric NMT domain adaptation, $k$-Nearest-Neighbor Machine Translation ($k$NN-MT) constructs an external datastore to store domain-specific translation knowledge, which derives a $k$NN distribution to interpolate the prediction distribution of the NMT model via a linear interpolation coefficient $\lambda$. 
Despite its success, $k$NN retrieval at each timestep leads to substantial time overhead. To address this issue, 
dominant studies resort to $k$NN-MT with adaptive retrieval ($k$NN-MT-AR), which dynamically estimates $\lambda$ and skips $k$NN retrieval if $\lambda$ is less than a fixed threshold. Unfortunately, $k$NN-MT-AR does not yield satisfactory results. In this paper, we first conduct a preliminary study to reveal two key limitations of $k$NN-MT-AR: 1) the optimization gap leads to inaccurate estimation of $\lambda$ for determining $k$NN retrieval skipping, and 2) using a fixed threshold fails to accommodate the dynamic demands for $k$NN retrieval at different timesteps. 
To mitigate these limitations, we then propose $k$NN-MT with dynamic retrieval ($k$NN-MT-DR) that significantly extends vanilla $k$NN-MT in two aspects. 
Firstly, 
we equip $k$NN-MT with a MLP-based classifier for determining whether to skip $k$NN retrieval at each timestep. 
Particularly, we explore several carefully-designed scalar features to fully exert the potential of the classifier. 
Secondly, we propose a timestep-aware threshold adjustment method to dynamically generate the threshold, which further improves the efficiency of our model.
Experimental results on the widely-used datasets demonstrate the effectiveness and generality of our model.\footnote{Our code is available at \url{https://github.com/DeepLearnXMU/knn-mt-dr}.
}
\end{abstract}
\section{Introduction}

As an effective paradigm for non-parametric domain adaptation, $k$-Nearest-Neighbor Machine Translation~($k$NN-MT)~\citep{khandelwal2020nearest} derives from $k$-Nearest-Neighbor Language Model~($k$NN-LM)~\citep{khandelwal2019generalization} and has garnered much attention recently~\citep{zheng2021adaptive, wang2022efficient, cao2023bridging, zhu2023ink}. Typically, $k$NN-MT introduces translation knowledge stored in an external datastore to enhance the NMT model, which can conveniently achieve non-parametric domain adaptation by changing external datastores.

In $k$NN-MT, a datastore containing key-value pairs is first constructed with an off-the-shelf NMT model, where the key is the decoder representation and the value corresponds to its target token. 
During translation, the current decoder representation is used as a query to retrieve $k$ nearest pairs from the datastore, where retrieved values are converted into a probability distribution. Finally, via a linear interpolation coefficient $\lambda$, this distribution is used to adjust the prediction distribution of the NMT model. In spite of success, retrieving at each timestep incurs substantial time overhead, which becomes considerable as the datastore expands.

To address this drawback, researchers have proposed 
two categories of approaches:
1)~$\emph{datastore compression}$ that improves retrieval efficiency by reducing the size of datastores~\citep{martins2022efficient,meng2022fast,wang2022efficient,dai2023simple,zhu2023knowledge,deguchi2023subset}; 2)~$\emph{retrieval reduction}$ that skips some $k$NN retrieval to speed up decoding. In this regard, the most representative work is $k$NN-MT with adaptive retrieval ($k$NN-MT-AR)~\citep{martins2022efficient} that skips $k$NN retrieval when the coefficient $\lambda$ is less than a fixed threshold $\alpha$. However, $k$NN-MT-AR does not achieve desired results as reported in~\citep{martins2022efficient}.

In this work, we mainly focus on the studies of $\emph{retrieval reduction}$, which is compatible with the other type of studies. To this end, we first re-implement $k$NN-MT-AR ~\citep{martins2022efficient} and conduct a preliminary study to analyze its limitations. Through in-depth analyses, we show that 
1) the optimization gap leads to inaccurate estimation of $\lambda$ for determining $k$NN retrieval skipping; 
2) with the increase in timesteps, the demand for $k$NN retrieval diminishes, which proves challenging for the fixed threshold $\alpha$ to handle effectively.

To overcome the above defects, we then significantly extend the vanilla $k$NN-MT into $k$NN-MT with dynamic retrieval ($k$NN-MT-DR), which accelerates the model decoding in two aspects. 
Concretely, instead of relying on the interpolation coefficient $\lambda$, we introduce a MLP-based classifier to explicitly determine whether to skip $k$NN retrieval as a binary classification task. Particularly, instead of using the decoder representation as the input of the classifier, we explore several carefully-designed scalar features to fully exert the potential of the classifier. 
Besides, we propose a timestep-aware threshold adjustment method to dynamically generate the threshold, so as to further improve the efficiency of our model.

To summarize, main contributions of our work include the following four aspects:
\begin{itemize}
\setlength{\itemsep}{2.5pt}
\setlength{\parsep}{2.5pt}
\setlength{\parskip}{1.5pt}
\item{Through in-depth analyses, we conclude two defects of $k$NN-MT-AR: the optimization gap leads to inaccurate estimation of $\lambda$ for $k$NN retrieval skipping, and a fixed threshold is unable to effectively handle the varying demands of $k$NN retrieval at different timesteps.}

\item{We propose to equip $k$NN-MT with an explicit classifier to determine whether to skip $k$NN retrieval, where carefully-designed features enable our model to achieve a better balance between model acceleration and performance.}

\item{We propose a timestep-aware threshold adjustment method to further improve the efficiency of our model.}

\item{Empirical evaluations on the multi-domain datasets validate the effectiveness of our model, as well as its compatibility with datastore compression methods.}

\end{itemize}

\section{Related Work}

\paragraph{Datastore Compression.} 
In this aspect, the size of the datastore for $k$NN retrieval is decreased to make retrieval efficient. For example, ~\citet{martins2022efficient} compress the datastore by greedily merging neighboring pairs that share the same values, and applying PCA algorithm~\citep{wold1987principal} to reduce the dimension of stored keys. Meanwhile, \citet{zhu2023knowledge} prune the datastore based on the concept of local correctness, while \citet{wang2022efficient} presents a cluster-based compact network to condense the dimension of stored keys, coupled with a cluster-based pruning strategy to discard redundant pairs. Additionally, some studies opt for dynamically adopting more compact datastores. For instance, for each token in the input sentence, \citet{meng2022fast} identify the relevant parallel sentences that contain this token and then collect corresponding word-aligned target tokens to construct a smaller datastore. Subsequently, \citet{dai2023simple} conduct sentence-level retrieval and dynamically construct a compact datastore for each input sentence. With the same motivation, \citet{deguchi2023subset} suggest retrieving target tokens from a subset of neighbor sentences related to the input sentence, where a look-up table based distance computation method is used to expedite retrieval.

\paragraph{Retrieval Reduction.} 
In this regard, some $k$NN retrieval is reduced to decrease time overhead for retrieval. For instance, ~\citet{martins2022chunk} adopt chunk-wise $k$NN retrieval rather than timestep-wise one, and~\citet{martins2022efficient} explore two approaches to reduce the frequency of $k$NN retrieval operations: 1) one introduces a caching mechanism to speed up decoding, where the cache mainly contains retrieved pairs from previous timesteps, and skip $k$NN retrieval if the distance between the query and any cached key is less than a predefined threshold; 2) the other proposes to conduct $k$NN retrieval when the interpolation coefficient $\lambda$ is less than a predefined threshold $\alpha$, which, however, does not achieve satisfactory results. 

Our work mainly focuses on the second type of studies mentioned above. We first conduct a preliminary study to in-depth analyze two limitations of the $\lambda$-based $k$NN retrieval skipping. To address these limitations, we introduce a classifier to explicitly determine whether to skip $k$NN retrieval as a classification task. Notably, almost concurrently with our work, \citet{shi2023towards} also use a classifier to speed up model decoding, sharing a similar motivation with ours. However, our work not only achieves better results, but also significantly differs from theirs in the following three aspects:

First, we explore several carefully-designed scalar features as the input for the classifier, which are crucial for achieving better performance. Second, when training the classifier, we adopt more reasonable criteria to construct training samples. To be specific, in addition to skipping retrieval when the target token ranks the $1$st position in the NMT prediction distribution, we believe that the model should also skip when the target token can not be obtained through $k$NN retrieval. Finally, based on the observation that the demand for $k$NN retrieval diminishes as timesteps increase, we propose a timestep-aware threshold method to further improve the efficiency of our model.

\section{Preliminary Study}
\subsection{Background}\label{intro_of_knn_mt}
Typically, given an off-the-shelf NMT model $\emph{f}_{\theta}$, a vanilla $k$NN-MT model is constructed through the following two stages:
\paragraph{Datastore Construction.} At this stage, all parallel sentence pairs in the training corpus $\mathcal{C}$=$\{({\bm{x}},{\bm{y}})\}$ are first fed into the NMT model $\emph{f}_{\theta}$ in a teacher-forcing manner~\citep{williams1989learning}. 
At each timestep $\emph{t}$, the decoder representation $h_t$ and its corresponding target token $y_t$ are collected to form a key-value pair, which is then added to the key-value datastore $\mathcal{D}{=}\{(h_t,y_t) \;|\; \forall y_t{\in}\bm{y} , (\bm{x},\bm{y})\}$, where $h_t$=$f_{\theta}(\bm{x},\bm{y}_{<t})$.
\paragraph{Translating with Retrieved Pairs.} During inference, the datastore is used to assist the NMT model. Specifically, the decoder representation $\hat{h}_{t}$ is used as a query to retrieve $k$ pairs $\mathcal{N}_{t}$=$\{({h}_{i},{y}_{i})\}_{i=1}^{k}$ from $\mathcal{D}$, which are then converted into a probability distribution over the vocabulary, abbreviated as $\emph{kNN distribution}$:
\begin{equation}
\begin{aligned}
p_{k{\rm NN}}(\hat{y_{t}}|\bm{x},\bm{y}_{<t})\propto \\ \sum_{({h}_{i},y_{i})\in\mathcal{N}_{t}}\mathds{1}_{(\hat{y}_{t}{=}{y}_{i})}&{\rm exp}({\frac{-\emph{d}(h_{i},\hat{h}_{t})}{\tau}}),
\end{aligned}
\end{equation}where $\mathds{1}_{(*)}$ is an indicator function, $\emph{d}(h_{i},\hat{h}_{t})$ measures the Euclidean distance between the query $\hat{h}_{t}$ and the key $h_{i}$, and $\tau$ is a predefined temperature. Finally, $k$NN-MT interpolates $p_{k{\rm NN}}$ with the prediction distribution $p_{\rm NMT}$ of the NMT model as a final translation distribution:
\begin{equation}
\begin{aligned}
p(\hat{y}_{t}|\bm{x},\bm{y}_{<t})=\lambda p_{k{\rm NN}} +(1{-}\lambda)p_{\rm NMT},
\end{aligned}
\end{equation}
where $\lambda$ denotes a predefined interpolation coefficient tuned on the validation set.

\paragraph{$k$NN-MT with Adaptive Retrieval} \label{analyses_of_$k$NNMT-AR}
Obviously, the retrieval of $k$NN-MT at each timestep incurs significant time overhead. To address this limitation, \citet{martins2022efficient}  follow \cite{he2021efficient} to explore $k$NN-MT with adaptive retrieval ($k$NN-MT-AR). Unlike the vanilla $k$NN-MT, they dynamically estimate the interpolation coefficient $\lambda$ using a light MLP network, which takes several neural and count-based features as the input. Then, they not only interpolate the $k$NN and NMT prediction distributions with $\lambda$, but also skip $k$NN retrieval when $\lambda$ is less than a fixed threshold $\alpha$. During training, they minimize the cross-entropy (CE) loss over the interpolated translation distribution.

Unfortunately, extensive results on several commonly-used datasets indicate that $k$NN-MT-AR does not achieve satisfactory results.
\begin{table}[!tbp]
	\renewcommand
	\arraystretch{1.2}
	\centering
\small
\begin{tabular}{c|ccccc}
\hline
$\boldsymbol{\alpha}$ & \bf{IT} & \bf{Koran} & \bf{Law} & \bf{Medical} & \bf{Subtitles} \\ \hline
$0.25$ & $0.27$ & $0.14$& $0.04$& $0.12$& $0.50$\\
$0.50$ & $0.50$ & $0.54$& $0.26$& $0.43$& $0.60$\\
$0.75$ & $0.51$ & $0.59$& $0.40$& $0.42$& $0.60$\\
\hline
\end{tabular}
\caption{
    F1 scores of the $\lambda$-based $k$NN
retrieval skipping of $k$NN-MT-AR~\citep{martins2022efficient} on the test sets.
	}\label{f1_scores_of_knn_retrieval}
\end{table}
\subsection{Limitations of $k$NN-MT-AR.}\label{analyses_of_lambda_based_method}
In this subsection, we conduct a preliminary study to explore the limitations of $k$NN-MT-AR. We strictly follow the settings of \citep{martins2022efficient} to re-implement their $k$NN-MT-AR, and then conduct two groups of experiments on the commonly-used multi-domain datasets released by~\citet{aharoni-goldberg-2020-unsupervised}.

As reported by \citet{martins2022efficient}, dynamically determining whether to skip $k$NN retrieval based on $\lambda$ leads to significant performance degradation. In the first group of experiments, to further provide evidence of this conclusion, we perform decoding on the test sets in a teacher-forcing manner and analyze the F1 scores of $\lambda$-based $k$NN retrieval skipping. 
As shown in Table~\ref{f1_scores_of_knn_retrieval}, 
F1 scores remain relatively low no matter which thresholds and datasets are used.

For the above results, 
we believe that there are two reasons leading to the inaccurate estimation of $\lambda$, which in turn makes $\lambda$ unsuitable for deciding whether to skip $k$NN retrieval. 

In addition to lacking the information of $k$NN distribution for $\lambda$ estimation\footnotemark, we believe that \textbf{the optimization objective of minimizing the CE loss over the translation distribution may be unsuitable to train an accurate} $\boldsymbol{\lambda}$ \textbf{estimator for determining $k$NN retrieval skipping}. 
\footnotetext{{Due to the consideration of model efficiency, $k$NN-MT-AR do not exploit the $k$NN retrieval information to estimate $\lambda$, which has been shown to be effective in previous studies~\citep{zheng2021adaptive,jiang-etal-2022-towards}.}}
To verify this claim, we consider whether to skip $k$NN retrieval as a standard binary classification task and use a binary CE loss to train a classifier for $\lambda$ estimation. Note that this classifier is also based on MLP and contains the same input as $k$NN-MT-AR. To avoid description confusion, we denote the $\lambda$ trained by $k$NN-MT-AR and the above binary CE loss as Tran-$\lambda$ and Bina-$\lambda$, respectively. 
Then, we calculate the average absolute value of the difference between Bina-$\lambda$ and Tran-$\lambda$ at all timesteps. The statistical results show that the average difference is $0.1495$, and $29.12$\% of timesteps exhibit a difference exceeding $0.2$.
These findings indicate significant differences between Bina-$\lambda$ and Tran-$\lambda$.

\begin{figure}[!t]

	\begin{center}
	\includegraphics[width=1.0\linewidth]{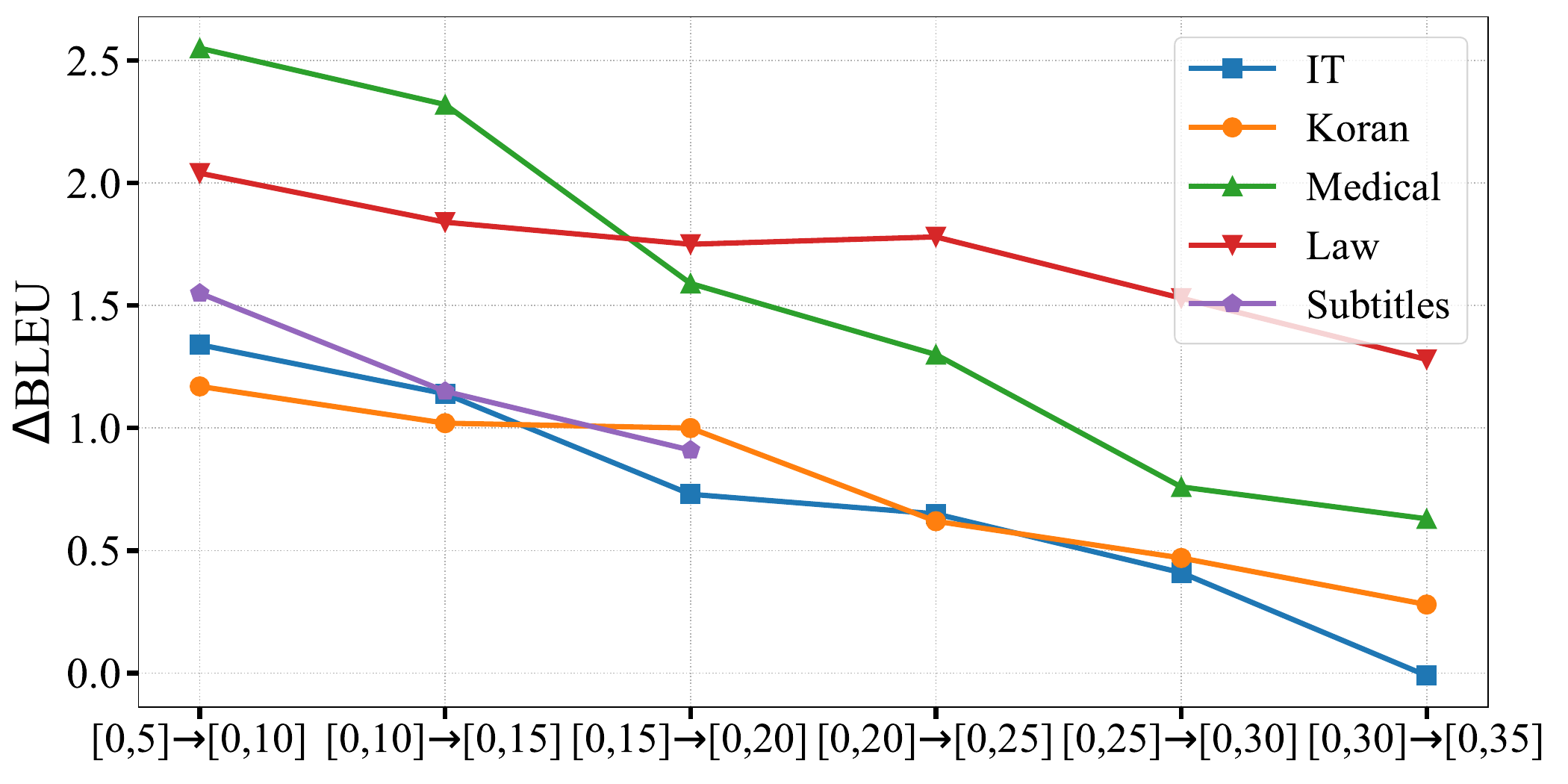}
	\end{center}
 	\caption{ The changes of BLEU improvements between adjacent intervals.
[0,5] means that $k$NN-MT only conducts retrieval when timestep ranges from 0 to 5. We only display the results for the first three BLEU improvements between adjacent intervals on the Subtitles, since the ratio of examples with length >= 25 is only about 1.35\%.
	}\label{delta_bleu}
\end{figure}

In the second group of experiments, we conduct experiments with vanilla $k$NN-MT on the validation sets to explore the impact of $k$NN retrieval during different timestep intervals. Specifically, we limit the model to only perform $k$NN retrieval in specific timestep intervals, where each interval starts from $0$ and increases by $5$ timesteps in length, and we only use instances with a translation length no less than the interval's right endpoint. From Figure~\ref{delta_bleu}, we observe that with the increase in timesteps, the performance gain caused by $k$NN retrieval gradually decreases across all datasets. 
This observation reveals that \textbf{the demand for $k$NN retrieval varies at different timesteps, which can not be handled well by the fixed threshold} $\boldsymbol{\alpha}$ \textbf{in $k$NN-MT-AR.}

In summary, the above two defects seriously limit the practicality of $k$NN-MT-AR.
Therefore, it is of great significance to explore more effective skipping $k$NN retrieval methods for $k$NN-MT.

\section{Our Model}
In this section, 
we significantly extend $k$NN-MT into $k$NN-MT-DR in the following two aspects.
\subsection{Classifier for Determining $k$NN Retrieval Skipping}\label{method_classifier}
Unlike $k$NN-MT-AR leveraging the interpolation coefficient $\lambda$ for determining whether to skip $k$NN retrieval, 
we directly equip $k$NN-MT with a binary classifier to determine whether to skip at each timestep. This classifier comprises a two-layer MLP network with ReLU activation. 
At timestep $t$, we conduct $k$NN retrieval only if the prediction probability of the classifier on conducting $k$NN retrieval exceeds a timestep-aware threshold $\alpha_{t}$, otherwise we will directly skip $k$NN retrieval. In the following, we will discuss the classifier, which involves the construction of training samples, input features, and the training objective.
\paragraph{Construction of Training Samples}\label{method_labels} 
To train the classifier, one crucial step is to construct training samples. 
In this regard, within the exploration of $k$NN-LM, 
\citet{he2021efficient} propose to construct training examples with two distinct labels, namely, ``conducting retrieval'' and ``skipping retrieval'', by comparing the prediction probabilities of $k$NN and NMT distributions on the target token $y_t$: 
when $p_{\rm kNN}(y_t)$ is greater than $p_{\rm NMT}(y_t)$, then $k$NN retrieval should be conducted, otherwise it can be skipped.
However, such a criterion still leads to a lot of redundant $k$NN retrieval. For example, when the target token $y_t$ has the highest probability in the NMT prediction distribution, there is no need to perform $k$NN retrieval, even if $p_{\rm kNN}(y_t){\ge}p_{\rm NMT}({y_t})$. Taking the IT validation set as an example, $69.8$\% of timesteps satisfy $p_{\rm kNN}(y_t){\ge}p_{\rm NMT}({y_t})$, among which $77.9$\% of the timesteps have $y_{t}$ ranking the $1$st position in the NMT prediction distribution.
Based on the above analysis, we traverse the parallel sentence pairs in the validation set, and collect various information at each timestep to construct training samples according to the following criteria:
\begin{itemize}
\setlength{\itemsep}{2.5pt}
\setlength{\parsep}{2.5pt}
\setlength{\parskip}{1.5pt}
\item{$k$NN retrieval should be skipped if one of the two conditions is satisfied: 1) $y_t$ ranks the $1$st position in the NMT prediction distribution, and 2) $y_t$ does not appear in the pairs obtained via $k$NN retrieval. Obviously, $k$NN retrieval yields no benefit in both conditions.}
\item{$k$NN retrieval should be conducted if $y_t$ is not the top-$1$ token in the NMT prediction distribution and it occurs in the $k$NN retrieval pairs. In this situation, conducting $k$NN distribution has the potential to improve translation.}
\end{itemize}

\paragraph{Input Features.} Unlike $k$NN-MT-AR, which uses the decoder representation and vectors mapped by other scalar features as the input, we consider several carefully-designed scalar features as the input for the classifier directly. By doing so, we reduce the input dimension, achieving effective training and enabling efficient inference. 
Here, we give detailed descriptions to these features:

\begin{table*}[!htbp]
	\renewcommand
	\arraystretch{1.2}
	\centering
\resizebox{1.0\linewidth}{!}{
\begin{tabular}{l|ccccc|c}
\hline
\bf{Model} & \bf{IT} & \bf{Koran} & \bf{Law} & \bf{Medical} & \bf{Subtitles} & \bf{Average} \\ 
\hline
Base NMT & $38.35$ / $82.74$ & $16.26$ / $72.04$ & $45.48$ / $85.66$ & $39.99$ / $83.13$ & $29.27$ / $79.76$ & $33.87$ / $80.67$\\
Vanilla $k$NN-MT & $45.83$ / $85.19$ & $20.37$ / $72.30$ & $61.16$ / $87.46$ & $54.22$ / $84.73$ & $31.28$ / $80.13$ & $42.57$ / $81.96$\\
\hline
$k$NN-MT-AR($\alpha$=$0.25$) & $43.20$ / $84.57$ & $19.57$ / $72.27$ & $59.89$ / $87.57$ & $53.12$ / $\bf{84.97}$ & $30.46$ / $80.04$ & $41.25$ / $\bf{81.88}$\\
$k$NN-MT-AR($\alpha$=$0.50$) & $41.19$ / $84.05$ & $17.23$ / $72.25$ & $58.83$ / $87.50$ & $51.22$ / $84.69$ & $29.45$ / $79.87$ & $39.58$ / $81.67$\\
$k$NN-MT-AR($\alpha$=$0.75$) & $39.05$ / $83.30$ & $16.40$ / $72.09$ & $51.11$ / $86.65$ & $45.14$ / $84.08$ & $29.30$ / $79.82$ & $36.20$ / $81.19$\\
Faster $k$NN-MT & $44.25$ / $84.59$ & $18.82$ / $72.07$ & $58.97$ / $87.36$ & $51.02$ / $84.45$ & $30.76$ / $80.04$ & $40.76$ / $81.70$\\
SK-MT$_1$ & $46.11$ / $84.39$ & $17.13$ / $72.16$ & $60.43$ / $87.46$ & $53.98$ / $84.22$ & $28.63$ / $77.52$ & $41.26$ / $81.15$\\
SK-MT$_2$ & $\bf{46.28}$ / $\bf{85.41}$ & $18.18$ / $72.17$ & $\bf{61.55}$ / $\bf{87.68}$ & $\bf{55.42}$ / $84.90$ & $28.14$ / $78.28$ & $\bf{41.91}$ / $81.69$\\
\rowcolor{lightgray}
Ours & $45.48$ / $84.60$ & $\bf{20.34}$ / $\bf{72.40}$ & $60.10$ / $87.39$ & $51.97$ / $84.36$ & $\bf{31.24}$ / $\bf{80.14}$ & $41.83$ / $81.78$\\
\hline
\end{tabular}
}
\caption{
    BLEU / COMET scores of various models on the multi-domain test sets.
	}\label{tab_main_performance}
\end{table*}

\begin{itemize}
\setlength{\itemsep}{2.5pt}
\setlength{\parsep}{2.5pt}
\setlength{\parskip}{1.5pt}
    \item {$p_{\rm NMT}(\hat{y}_t)$}: the probability of the top-$1$ predicted token $\hat{y}_t$ in the NMT prediction distribution. The higher the prediction confidence of the NMT model, the more likely the $\hat{y}_t$ to be the correct one. In this situation, $k$NN retrieval is more likely to be skipped.
    
    \item {$\| \hat{h}_{t} \|_2$}: the $L_2$ norm of current decoder representation. Inspired by~\citep{liu-etal-2020-norm}, we use the vector norm of the decoder representation $\hat{h}_{t}$ to measure the translation difficulty at current timestep: the larger $\| \hat{h}_{t} \|_2$, the more difficult the translation is.

    \item {$\text{max}(attn)$}: the maximal weight of the cross-attention in the last layer of the decoder during current decoding timestep. A large weight means that the NMT model is relatively certain about which source token to be translated. In this case, the translation difficulty is often relatively low.

\end{itemize}
Finally, these features are concatenated and normalized with batch normalization~\citep{ioffe2015batch} before being the input for the classifier.
\paragraph{Classifier Training.} To achieve efficient domain adaptation for NMT, we fix the parameters of NMT model and only update those of classifier during training. Following \citet{he2021efficient}, we select $90$\% of the validation set to train the classifier, and use the remaining $10$\% for validation. Then, according to the above criterion, we construct training samples with different labels at each timestep to train our classifier. Considering the significant imbalance between two classes of training samples\footnote{Through data analysis, we find that only $16.8$\% of training samples require $k$NN retrieval in the IT validation set.}, we adopt Focal Loss~\citep{lin2017focal} to train our classifier as follows:
\begin{equation}\label{eq_focal_loss}
\begin{aligned}
\mathcal{L}(p_c)=-\alpha_c(1-p_c)^\gamma{log(p_c)},
\end{aligned}
\end{equation}where $c$$=$$0$$/$$1$ denotes the label of skipping/conducting $k$NN retrieval, $p_c$ is the prediction probability of the classifier on the label $c$, $\alpha_c$ is a weighting factor controlling the balance between different kinds of samples, and $\gamma$ is a hyper-parameter adjusting the impacts of loss functions of easy and hard samples~\citep{lin2017focal}.
\subsection{Timestep-aware Threshold Adjustment}\label{method_threshold}
As analyzed in Section~\ref{analyses_of_lambda_based_method}, the benefit of $k$NN retrieval diminishes with the increase in timesteps, indicating that using the fixed threshold $\alpha$ is not the most reasonable choice. 
To deal with this issue, we propose a timestep-aware threshold adjustment method to accommodate the varied demands of $k$NN retrieval. Formally, we heuristically define a dynamic threshold function specific to the timestep:
\begin{equation}
\begin{aligned}
\alpha_t = \alpha_{\text{min}} + \operatorname{clip}(\frac{t}{T}; 0, 1)^2\times(0.5 - \alpha_{\text{min}})
\end{aligned}
\end{equation}

where $\operatorname{clip}(x; a, b)$ clamp $x$ within the range of $[a, b]$, $t$ is the decoding timestep, $\alpha_{\text{min}}$ is the lower limit of threshold, and $T$ is the average length of sentences in the validation set. Apparently, with the increase of $t$, $\alpha_t$ will gradually increase until it reaches $0.5$.

\section{Experiments}
\subsection{Setup}
\paragraph{Datasets.} We conduct experiments on the multi-domains dataset released by~\citet{aharoni-goldberg-2020-unsupervised}. The dataset comprises German-English parallel corpora across five domains: Koran, IT, Medical, Law, and Subtitles, the detailed statistics can be found in Appendix~\ref{dataset_statistics}. We employ Byte Pair Encoding~\citep{DBLP:conf/acl/SennrichHB16a} to split words into subwords. Finally, we use two metrics to evaluate the translation quality: SacreBLEU\footnote{\url{https://github.com/mjpost/sacrebleu}}~\citep{post-2018-call} and COMET\footnote{\url{https://github.com/unbabel/COMET}}~\citep{DBLP:conf/emnlp/ReiSFL20}.
\paragraph{Model Configuration.} We develop our model with $k$NN-BOX\footnote{\url{https://github.com/NJUNLP/knn-box}}~\citep{zhu2023knn} and use Faiss~\citep{johnson2019billion} to build the datastore and search nearest neighbors. 
To ensure fair comparisons, we adopt the same settings as the previous study~\citep{khandelwal2020nearest}. Concretely, we set the number of retrieved pairs to 8, the temperature $\tau$ to $100$ for Koran and $10$ for the other datasets, and $\lambda$ to $0.7$ for IT, Subtitles, $0.8$ for the other datasets. We use a two-layer MLP network with ReLU activation~\citep{agarap2018deep} to construct our classifier, of which hidden size is set to $32$ because it is not sensitive in our model. Besides, we set the hyper-parameter $\alpha_{\text{min}}$ to $0.45$ for Koran, Subtitles, $0.4$ for the other datasets.\footnote{The details of tuning $\alpha_{\text{min}}$ are reported in Appendix~\ref{hyper-parameter_tuning}.}

\paragraph{Baselines.} Our baselines include:
\begin{itemize}
\setlength{\itemsep}{2.5pt}
\setlength{\parsep}{2.5pt}
\setlength{\parskip}{1.5pt}
    \item{{\bf Base NMT}}~\citep{ng-etal-2019-facebook}. Following~\citet{khandelwal2020nearest}, we use the WMT’19 German-English news translation task winner as the base NMT model.
    \item{{\bf Vanilla $k$NN-MT}}~\citep{khandelwal2020nearest}. It serves as a baseline, upon which we develop our model.
    \item{{\bf $k$NN-MT-AR }}~\citep{martins2022efficient}. It performs retrieval only when the interpolation coefficient $\lambda$ is less than a predefined threshold~$\alpha$. Note that it is our most important baseline. Particularly, we report the performance of $k$NN-MT-AR with $\alpha$ set to $0.25$, $0.50$, and $0.75$, respectively.
    \item{{\bf Faster $k$NN-MT}}~\citep{shi2023towards}. It is a concurrent work with ours, where a two-layer MLP network takes decoder representation as the input to determine whether to skip $k$NN retrieval at each timestep.
    \item{{\bf SK-MT}}~\citep{dai2023simple}. It dynamically constructs a compact datastore by conducting sentence-level retrieval for each input sentence. Specially, we report the performance of SK-MT$_1$ with $m=2$, $k=1$ and SK-MT$_2$ with $m=16$, $k=2$.
    
    \setlength{\belowcaptionskip}{0. cm}
    \setlength{\abovecaptionskip}{0. cm}
\end{itemize}

\begin{table}[!t]
	\renewcommand
	\arraystretch{1.15}
	\centering
\resizebox{1.0\linewidth}{!}{
\begin{tabular}{l|ccccc}
\hline
\bf{Model} & \bf{IT} & \bf{Koran} & \bf{Law} & \bf{Medical} & \bf{Subtitles} \\ \hline
\multicolumn{6}{c}{\bf{Batch Size = 128}} \\ \hline
Base NMT & $3270.84$& $3912.95$& $3690.85$& $3152.59$& $4004.40$\\
Vanilla $k$NN-MT & $2584.31$ & $3287.24$ & $2300.23$ & $2363.00$& $478.99$\\
\hline
$k$NN-MT-AR  & $2724.76$& $3069.38$& $2241.93$& $2382.52$& $886.16$\\
Faster $k$NN-MT & $2912.67$ & $\bf{3609.53}$ & $2923.79$ & $\bf{2676.11}$ & $999.57$ \\
SK-MT$_1$ & $524.65$ & $537.06$ & $533.52$ & $560.14$ & $264.30$ \\
SK-MT$_2$ & $385.95$ & $408.21$ & $423.16$ & $428.63$ & $236.42$ \\
\rowcolor{lightgray}Ours & $\bf{2944.38}$ & $3522.49$ & $\bf{2933.76}$ & $2605.12$ & $\bf{1002.13}$ \\
\hline
\multicolumn{6}{c}{\bf{Batch Size = 64}} \\
\hline
Base NMT & $3150.95$ & $3730.90$ & $3607.41$  & $3111.54$  & $3377.17$ \\
Vanilla $k$NN-MT & $2506.85$ & $2945.54$ & $2252.18$ & $2329.36$& $445.88$ \\
\hline
$k$NN-MT-AR  & $2789.59$& $2678.89$& $2125.88$& $2323.60$ & $794.04$\\
Faster $k$NN-MT & $2783.62$ & $3124.68$ & $2726.92$ & $\bf{2592.75}$ & $898.26$ \\
SK-MT$_1$ & $518.82$&$525.16$&$524.28$&$547.08$&$258.02$ \\
SK-MT$_2$ & $381.87$ & $396.00$ & $411.91$ & $420.05$ & $224.41$ \\
\rowcolor{lightgray}Ours & $\bf{2798.39}$ & $\bf{3132.26}$ & $\bf{2755.02}$  & $2575.40$ & $\bf{901.33}$\\
\hline
\multicolumn{6}{c}{\bf{Batch Size = 32}} \\
\hline
Base NMT & $2559.84$ & $2933.82$ & $2995.43$ & $2688.93$ & $2635.05$ \\
Vanilla $k$NN-MT & $2001.80$ & $2360.50$& $1908.76$ & $1955.65$& $408.54$\\
\hline
$k$NN-MT-AR  & $2067.55$ & $1792.74$ & $1925.76$&  $1694.34$& $676.28$\\
Faster $k$NN-MT & $\bf{2131.48}$ & $\bf{2432.76}$ & $2225.68$ & $\bf{2047.19}$ & $735.26$ \\
SK-MT$_1$ & $486.17$&$500.06$&$494.30$&$523.16$&$247.17$ \\
SK-MT$_2$ & $360.76$&$374.32$&$392.97$&$400.63$&$203.85$ \\
\rowcolor{lightgray}Ours & $2117.94$ & $2392.60$ & $\bf{2226.63}$ & $2031.51$ & $\bf{737.85}$\\
\hline
\multicolumn{6}{c}{\bf{Batch Size = 16}} \\
\hline
Base NMT & $1577.03$ & $1878.36$ & $1959.55$ & $1737.23$ & $1686.02$ \\
Vanilla $k$NN-MT & $1378.65$ & $1429.78$ & $1318.55$ & $1366.35$ & $340.96$ \\
\hline
$k$NN-MT-AR  & $1369.49$ & $1437.82$ & $1244.21$  & $1323.07$ & $506.17$ \\
Faster $k$NN-MT & $1396.32$ & $1451.95$ & $1455.46$ & $\bf{1406.26}$ & $538.65$ \\
SK-MT$_1$ & $410.57$&$409.76$&$431.25$&$440.31$&$220.65$ \\
SK-MT$_2$ & $318.62$&$340.16$&$354.70$&$355.37$&$176.17$ \\
\rowcolor{lightgray}
Ours & $\bf{1441.66}$ & $\bf{1487.54}$ & $\bf{1472.04}$ & $1395.21$ & $\bf{546.22}$ \\
\hline
\multicolumn{6}{c}{\bf{Batch Size = 1}} \\
\hline
Base NMT & $159.24$ & $168.84$ & $173.22$ & $171.12$ & $159.04$ \\
Vanilla $k$NN-MT & $136.19$ & $139.02$ & $142.91$ & $138.31$ & $42.75$\\
\hline
$k$NN-MT-AR & $127.23$ & $130.35$ & $127.93$ & $128.09$ & $57.98$ \\
Faster $k$NN-MT & $139.54$ & $\bf{140.85}$ & $147.18$ & $\bf{140.68}$ & $58.46$ \\
SK-MT$_1$ & $89.76$&$103.97$&$96.42$&$92.52$&$35.26$ \\
SK-MT$_2$ & $84.10$&$97.01$&$89.82$&$85.72$&$32.68$ \\
\rowcolor{lightgray}Ours & $\bf{139.84}$ & $140.18$ & $\bf{147.44}$  & $139.84$  & $\bf{58.62}$ \\
\hline
\end{tabular}
}
\caption{
    Decoding speed (\#Tok/Sec$\uparrow$) of various models using different batch sizes on the multi-domain test sets. Here, we only display the decoding speed of $k$NN-MT-AR($\alpha$=0.25), since $k$NN-MT-AR($\alpha$=0.5) and $k$NN-MT-AR($\alpha$=0.75) exhibit significant performance degradation, as reported in Table~\ref{tab_main_performance}. 
    All results are evaluated on 
    an NVIDIA RTX A6000 GPU.
	}\label{tab_main_speed}
\end{table}

\begin{table}[!t]
	\renewcommand
	\arraystretch{1.2}
	\centering
\small
\begin{tabular}{l|c}
\hline
\bf{Model} & \bf{BLEU} \\ 
\hline
Faster $k$NN-MT & $44.25$\\
\hdashline
Ours & $45.48$  \\
\quad \emph{Our Criteria}$\Rightarrow$\emph{Conventional Criteria} & $43.90$ \\
\quad \emph{Dynamic Threshold}$\Rightarrow$\emph{Fixed Threshold} & $44.28$ \\
\quad \emph{Focal Loss}$\Rightarrow$\emph{Weighted CE Loss} & $44.79$ \\
\hdashline
\quad {\emph{w/o} $p_{\rm NMT}(\hat{y}_t)$} & $44.62$  \\
\quad {\emph{w/o} $\| \hat{h}_{t} \|_2$} & $45.01$ \\
\quad {\emph{w/o} $\text{max}(Attn)$} & $45.12$ \\

\hline
    \end{tabular}
\caption{
    Ablation studies on the IT test set. 
	}\label{tab_ablation}
\end{table}

\subsection{Main Results}
To comprehensively evaluate various models, we report their translation quality and decoding speed.
\paragraph{Translation Quality.} Table~\ref{tab_main_performance} presents BLEU and COMET scores of various models on the multi-domain test sets. We observe that  both $k$NN-MT-AR and Faster $k$NN-MT suffer from significant performance declines compared to Vanilla $k$NN-MT, echoing with the results reported in previous studies~\citep{martins2022efficient,shi2023towards}. 
In contrast, 
our model exhibits the least performance degradation.
Specifically, our model achieves average BLEU and COMET scores of $41.83$ and $81.78$ points, with only $0.74$ and $0.18$ points lower than those of Vanilla $k$NN-MT, respectively. Although SK-MT$_2$ performs better than our model, experiments in Section~\ref{experiments_on_adaptive_kNN-MT} find that it is not compatible with Adaptive $k$NN-MT, while our model significantly outperforms SK-MT$_2$ when using Adaptive $k$NN-MT as the base model.
\paragraph{Decoding Speed.} \label{analyses_of_decoding_speeds}
Model efficiency is a crucial performance indicator for $k$NN-MT. 
As implemented in previous studies \citep{zheng2021adaptive,deguchi2023subset}, we try different batch sizes: $1$, $16$, $32$, $64$ and $128$, and then report the model efficiency using ``$\emph{\#Tok/Sec}$'': the number of translation tokens generated by the model per second.

Experimental results are listed in Table~\ref{tab_main_speed}. We have the following interesting findings: First, regardless of the batch size used, our model is more efficient than both Vanilla $k$NN-MT, $k$NN-MT-AR($\alpha$=$0.25$), SK-MT$_1$ and SK-MT$_2$. 

Second, as the batch size increases, the efficiency advantage of our model becomes more apparent. On most datasets, we find that the acceleration ratios of our model with large batch sizes (64 or 128) are significantly higher than those with small batch sizes~(1 or 16).
Finally, with the increase of the datastore size, the efficiency advantage of our model also becomes more significant. As analysed in Appendix~\ref{dataset_statistics}, the datastore in Subtitles contains the maximum number of pairs while the datastore in Koran is the smallest. Correspondingly, our model has the most significant acceleration effect on the Subtitles dataset, while the acceleration effect on the Koran dataset is the least significant.

Based on the above experimental results, we believe that compared with baselines, ours can achieve better balance between model performance degradation and acceleration.

\begin{table*}[!htbp]
	\renewcommand
	\arraystretch{1.2}
	\centering
\resizebox{1.0\linewidth}{!}{
\begin{tabular}{l|ccccc|c}
\hline
\bf{Model} & \bf{IT} & \bf{Koran} & \bf{Law} & \bf{Medical} & \bf{Subtitles} & \bf{Average} \\ 
\hline
SK-MT$_1$ & $46.11$ / $84.39$ & $17.13$ / $72.16$ & $60.43$ / $87.46$ & $53.98$ / $84.22$ & $28.63$ / $77.52$ & $41.26$ / $81.15$\\
SK-MT$_2$ & $46.28$ / $85.41$ & $18.18$ / $72.17$ & $61.55$ / $87.68$ & $55.42$ / $84.90$ & $28.14$ / $78.28$ & $41.91$ / $81.69$\\
Adaptive $k$NN-MT & $47.26$ / $85.99$ & $20.15$ / $73.22$ & $62.68$ / $88.07$ & $56.49$ / $85.25$ & $31.49$ / $80.25$ & $43.61$ / $82.56$\\
\quad + $k$NN-MT-AR($\alpha$=$0.25$)  & $44.34$ / $84.92$ & $\bf{20.19}$ / $72.40$ & $\bf{61.86}$ / $87.66$ & $\bf{55.46}$ / $84.76$ & $30.64$ / $79.92$ & $42.50$ / $81.93$\\
\quad + $k$NN-MT-AR($\alpha$=$0.50$)  & $41.34$ / $84.51$ & $17.04$ / $72.05$ & $59.71$ / $87.37$ & $52.33$ / $84.59$ & $29.37$ / $79.83$ & $39.96$ / $81.67$\\
\quad + $k$NN-MT-AR($\alpha$=$0.75$)  & $39.22$ / $83.69$ & $16.48$ / $72.06$ & $51.28$ / $86.60$ & $45.23$ / $84.08$ & $29.30$ / $79.81$ & $36.30$ / $81.25$\\
\quad + Faster $k$NN-MT & $45.38$ / $85.43$ & $19.04$ / $72.98$ & $59.95$ / $87.73$ & $53.09$ / $84.91$ & $30.63$ / $80.06$ & $41.62$ / $82.22$\\
\rowcolor{lightgray}
\quad + Ours & $\bf{46.94}$ / $\bf{85.46}$ & $20.05$ / $\bf{73.26}$ & $61.17$ / $\bf{87.75}$ & $54.58$ / $\bf{84.98}$ & $\bf{31.35}$ / $\bf{80.38}$ & $\bf{42.82}$ / $\bf{82.37}$\\
\hline
\end{tabular}
}
\caption{
    BLEU / COMET scores of various models based on Adaptive $k$NN-MT. 
	}\label{tab_adaptive_performance}
\end{table*}
\begin{table*}[!htbp]
	\renewcommand
	\arraystretch{1.2}
	\centering
\small
\begin{tabular}{l|ccccc}
\hline
\bf{Model} & \bf{IT} & \bf{Koran} & \bf{Law} & \bf{Medical} & \bf{Subtitles} \\ \hline
SK-MT$_1$ & $524.65$ & $537.06$ & $533.52$ & $560.14$ & $264.30$ \\
SK-MT$_2$ & $385.95$ & $408.21$ & $423.16$ & $428.63$ & $236.42$ \\
Adaptive $k$NN-MT & $2583.92$& $3320.01$& $2292.75$& $2368.51$& $484.62$\\
\quad + $k$NN-MT-AR($\alpha$=$0.25$) & $2646.95$& $3098.34$& $2191.50$& $2235.01$& $873.98$\\
\quad + Faster $k$NN-MT & $2923.62$ & $\bf{3665.24}$ & $\bf{2901.53}$ & $\bf{2733.55}$& $952.27$\\
\rowcolor{lightgray}
\quad + Ours & $\bf{2971.77}$& $3569.44$ & $2883.89$& $2712.45$& $\bf{1075.36}$\\
\hline
\end{tabular}
\caption{
   Decoding speed (\#Tok/Sec$\uparrow$) of various models based on Adaptive $k$NN-MT. Note that we also omit the results of $k$NN-MT-AR($\alpha$=0.5) and $k$NN-MT-AR($\alpha$=0.75). Here, we set the batch size as $128$.
	}\label{tab_adaptive_speeds}
\end{table*}

\begin{table*}[!t]
	\renewcommand
	\arraystretch{1.2}
	\centering
\small
\begin{tabular}{l|ccccc|c}
\hline
\bf{Model} & \bf{IT} & \bf{Koran} & \bf{Law} & \bf{Medical} & \bf{Subtitles} & \bf{Average}\\ \hline
PLAC & $46.81$ / $85.65$ & $20.51$ / $73.21$ & $62.89$ / $88.01$ & $56.05$ / $85.16$ & $31.59$ / $80.36$ & $43.57$ / $82.48$\\
\quad + Ours & $46.83$ / $85.40$ & $20.36$ / $73.25$ & $61.66$ / $87.82$ & $54.82$ / $85.01$ & $31.28$ / $80.29$ & $42.99$ / $82.35$\\
\hline
PCK & $47.27$ / $86.43$ & $19.93$ / $72.96$ & $62.91$ / $88.03$ & $56.46$ / $85.15$ & $31.69$ / $80.53$ & $43.65$ / $82.62$\\
\quad + Ours & $46.85$ / $85.97$ & $19.99$ / $73.24$ & $61.98$ / $88.05$ & $55.34$ / $85.11$ & $31.20$ / $80.44$ & $43.07$ / $82.56$\\
\hline
\end{tabular}
\caption{
   BLEU / COMET scores of PLAC~\citep{zhu2023knowledge} and PCK~\citep{wang2022efficient}, alongside these integrated with ours. 
	}\label{tab_main_comp_scores}
\end{table*}

\begin{table}[!t]
	\renewcommand
	\arraystretch{1.2}
	\centering
\resizebox{1.05\linewidth}{!}{
\begin{tabular}{l|ccccc}
\hline
\bf{Model} & \bf{IT} & \bf{Koran} & \bf{Law} & \bf{Medical} & \bf{Subtitles} \\ \hline
PLAC & $2684.36$ & $3398.53$ & $2433.44$ & $2383.00$& $749.49$\\
\ +Ours & $3027.95$ & $3596.20$ & $3025.14$  & $2713.74$  & $1461.30$ \\
\hline
PCK & $2873.40$ & $3535.19$ & $2673.76$ & $2617.73$& $979.52$\\
\ +Ours & $3072.21$ & $3588.76$ & $3009.64$& $2720.04$ & $1801.97$ \\
\hline
\end{tabular}
}
\caption{
   Decoding speed (\#Tok/Sec$\uparrow$) of PLAC~\citep{zhu2023knowledge} and PCK~\citep{wang2022efficient}, alongside these integrated with ours. Here, we set the batch size as $128$.
	}\label{tab_main_comp_speeds}
\end{table}

\subsection{Ablation Studies}
Following previous studies \cite{zheng2021adaptive, jiang-etal-2022-towards}, 
we compare our model with its variants on the IT test set. 
As shown in Table~\ref{tab_ablation}, we consider the following variants:
\begin{itemize}
\setlength{\itemsep}{2.5pt}
\setlength{\parsep}{1.5pt}
\setlength{\parskip}{1.5pt}
\item{\emph{Our Criteria}$\Rightarrow$\emph{Conventional Criteria.}} As mentioned in Section~\ref{method_classifier}, we adopt new criteria to determine whether $k$NN retrieval in training samples can be skipped.
To verify the effectiveness of our criteria, we compare our criteria with the conventional criteria as mentioned in~\citet{he2021efficient}: the $k$NN retrieval should be conducted if $p_{\rm kNN}(y_t){\geq}p_{\rm NMT}({y_t})$, otherwise it can be skipped. 
We first report the proportion changes between two labels of training samples on the IT dataset. Using the conventional criteria, the proportion of training samples labeled as skipping retrieval is about $30.2$\%, which is significantly smaller than the proportion $83.2$\% in our criteria. Obviously, more $k$NN retrieval can be skipped with our criteria. Second, we focus on the change of model performance. From Line $2$, we observe that the conventional criteria leads to a significant performance degeneration, which strongly reveals the effectiveness of our critera.
\item{\emph{Dynamic Threshold}$\Rightarrow$\emph{Fixed Threshold.}} We replace the proposed dynamic threshold $\alpha_{t}$ mentioned in Section \ref{method_threshold} with the originally-used fixed threshold $\alpha$$=$$0.5$ in this variant. As shown in Line $3$, we observe that removing the dynamic threshold leads to a performance decline, demonstrating the effectiveness of our threshold adjustment method.
\item{\emph{Focal Loss}$\Rightarrow$\emph{Weighted CE Loss}.} To make a fair comparison, we follow \citet{shi2023towards} to adopt a weighted CE loss, which sets $\gamma$ as $0$ in Equation~\ref{eq_focal_loss}. Back to Table \ref{tab_ablation}, we find that this variant is inferior to our model in terms of translation quality. However, it still surpasses Faster $k$NN-MT with a large margin, confirming the significant advantage of our model in translation quality.

\item{\emph{w/o Input Features.}} To verify the benefit of our carefully-designed features, we thoroughly construct several variants, each of which discards one kind of feature to train the classifier. As shown in Lines $6$-$8$, all variants exhibit performance drops with varying degrees. Thus, we confirm all features are useful for our classifier. 
\end{itemize}

\begin{table*}[!h]
	\renewcommand
	\arraystretch{1.2}
	\centering
\resizebox{1.0\linewidth}{!}{
\begin{tabular}{l|ccccc|c}
\hline
\bf{Model} & \bf{IT} & \bf{Koran} & \bf{Law} & \bf{Medical} & \bf{Subtitles} \\ 
\hline
Vanilla $k$NN-MT & $45.72$ / $467.21$ & $19.38$ / $534.79$ & $61.22$ / $456.88$ & $54.11$ / $501.02$ & $31.62$ / $515.47$\\
\hline
$k$NN-MT-AR($\alpha$=$0.25$) & $43.56$ / $569.73$ & $19.10$ / $598.69$ & $59.42$ / $533.70$ & $51.20$ / $530.95$ & $30.78$ / $634.12$\\
Faster $k$NN-MT & $43.79$ / $762.51$ & $17.82$ / $\bf{1108.25}$ & $58.82$ / $\bf{1155.10}$ & $50.51$ / $\bf{1076.66}$ & $30.71$ / $1048.35$ \\
SK-MT$_1$ & $45.36$ / $306.53$ & $16.24$ / $236.57$ & $60.21$ / $310.59$ & $53.78$ / $346.35$ & $26.87$ / $265.72$ \\
SK-MT$_2$ & $\bf{45.51}$ / $258.14$ & $17.12$ / $184.36$ & $\bf{60.62}$ / $277.27$ & $\bf{55.10}$ / $277.27$ & $28.40$ / $214.33$ \\
\rowcolor{lightgray}
Ours & $45.24$ / $\bf{886.54}$ & $\bf{19.17}$ / $880.50$& $60.23$ / $949.79$& $52.59$ / $1040.25$& $\bf{31.12}$ / $\bf{1078.92}$\\
\hline
\end{tabular}
}
\caption{
    BLEU$\uparrow$ and \#Tok/Sec$\uparrow$ of various models on the all-domain datastore.
	}\label{tab_all_domain}
\end{table*}

\subsection{Experiments on Adaptive $k$NN-MT}
\label{experiments_on_adaptive_kNN-MT}
Adaptive $k$NN-MT~\citep{zheng2021adaptive} is a widely-used variant of $k$NN-MT and significantly outperforms Vanilla $k$NN-MT in terms of performance. It introduces a meta-$k$ network, a two-layer MLP incorporating distances and counts of all $k$NN retrieval pairs, to dynamically estimate $\lambda$. 
Our model can also utilize Adaptive $k$NN-MT as the base models. When using Adaptive $k$NN-MT as the base model, we dynamically estimate $\lambda$ solely for timesteps considered to conduct $k$NN retrieval. Additionally, we explore the performance of Adaptive $k$NN-MT as the base model for $k$NN-MT-AR. To ensure fairness, we employ the $\lambda$ of $k$NN-MT-AR to determine whether to skip $k$NN retrieval, and interpolate using the $\lambda$ of Adaptive $k$NN-MT.

We also report the translation quality and decoding speed, as shown in the Table~\ref{tab_adaptive_performance} and Table~\ref{tab_adaptive_speeds}, respectively. Our model also demonstrate the least performance decline and achieve the most efficient decoding speed. Although Faster $k$NN-MT demonstrates comparable decoding speeds to ours, our model achieves superior performance.

\subsection{Compatibility with Datastore Compression Methods}

In this group of experiments,
we choose PLAC~\citep{zhu2023knowledge} and PCK~\citep{wang2022efficient} as the basic models for our compatibility experiment, both of which are derived from Adaptive $k$NN-MT. Typically, PLAC prunes the datastore by eliminating pairs with high knowledge margin values, while PCK introduces a cluster-based compact network to condense the dimension of stored keys and utilizes a cluster-based pruning strategy to discard redundant pairs.

Tables \ref{tab_main_comp_scores} and \ref{tab_main_comp_speeds} report the translation quality and decoding speed, respectively.
We can observe that our model can further improve the efficiency of these two models, with slight drops in translation quality. Thus, we confirm that ours is also compatible with both PLAC and PCK.

\subsection{All-Domains Datastore Experiment}
\label{All-Domains Datastore}

To provide more evidences for the efficiency of our model, we follow \citet{khandelwal2020nearest} to conduct the experiment on the all domains datastore. We report the BLEU scores and decoding speed as shown in Table~\ref{tab_all_domain}. Although SK-MT$_2$ significantly outperforms ours in the medical domain, it exhibits a significant slowdown in decoding speed across all domains. In contrast, our model achieves the best balance between translation quality and decoding speed.

\subsection{Evaluation on Other Languages}
\label{Zh-En Experiments}

In order to further validate the generality of our model, we adopt the same settings as the previous study \cite{zhu2023knowledge} to perform  experiments on Chinese-to-English translation using the Laws and Thesis domains from the UM dataset\cite{tian2014corpus}. As reported in Table~\ref{tab_zh_en}, it is observable that ours achieves a more efficient decoding speed with almost no loss in performance.

\begin{table}[!h]
	\renewcommand
	\arraystretch{1.2}
	\centering
\resizebox{1.0\linewidth}{!}{
\begin{tabular}{l|ccccc|c}
\hline
\bf{Model} & \bf{Laws} & \bf{Thesis} \\ 
\hline
Base NMT & $14.48$ / $5578.24$ & $12.23$ / $5985.98$\\
Adaptive $k$NN-MT & $31.61$ / $3142.54$ & $15.96$ / $3389.67$\\
\hline
$k$NN-MT-AR($\alpha$=$0.25$) & $27.66$ / $3233.05$ & $13.54$ / $3555.49$\\
Faster $k$NN-MT & $27.86$ / $\bf{3619.14}$ & $13.45$ / $3882.05$\\
SK-MT$_1$ & $27.02$ / $604.16$ & $15.18$ / $589.71$ \\
SK-MT$_2$ & $27.21$ / $547.20$ & $15.33$ / $564.37$ \\
\rowcolor{lightgray}
Ours & $\bf{31.72}$ / $3457.92$ & $\bf{15.83}$ / $\bf{3989.80}$\\
\hline
\end{tabular}
}
\caption{
    BLEU$\uparrow$ and \#Tok/Sec$\uparrow$ of various models on the UM dataset.
	}\label{tab_zh_en}
\end{table}

\section{Conclusion and Future Work}
In this work, we first in-depth analyze the limitations of $k$NN-MT-AR, and then significantly extend the vanilla $k$NN-MT to $k$NN-MT-DR in two aspects. First, we equip the model with a classifier to determine whether to skip $k$NN retrieval, where several carefully-designed scalar features are exploited to exert the potential of the classifier. Second, we propose a timestep-aware threshold adjustment method to further refine $k$NN retrieval skipping. Extensive experiments and analyses verify the effectiveness of our model.

Inspired by \citep{li-etal-2023-revisiting}, we will further improve our model by incorporating more source-side information into our classifier. 
Besides, we aim to generalize our model to $k$NN-LM~\citep{khandelwal2019generalization} and multilingual scenario~\citep{stap-monz-2023-multilingual}, so as to validate its generalizability.

\section*{Limitations}
As our model integrates an additional classifier, there is an associated increase in time consumption. Notably, as the size of the datastore decreases, the time overhead for $k$NN retrieval diminishes and classifier-related time cost becomes more apparent, which results in a less pronounced acceleration in decoding. Besides, the experiments of decoding speed are evaluated solely on a single computer, while the time overhead of $k$NN retrieval may differ across different hardware, yielding varied acceleration results.

\section*{Acknowledgements}
The project was supported by National Natural Science Foundation of China (No. 62036004, No. 62276219), and the Public Technology Service Platform Project of Xiamen (No. 3502Z20231043).
We also thank the reviewers for their insightful comments.

\bibliography{custom}

\appendix

\section{Dataset Statistics}
\label{dataset_statistics}

The number of parallel sentence pairs in different datasets and the sizes of the constructed datastores are shown in Table~\ref{tab_stat}.

\begin{table}[!h]
	\renewcommand
	\arraystretch{1.15}
	\centering
\resizebox{1.0\linewidth}{!}{
\begin{tabular}{c|ccccc}
\hline
\bf{Dataset} & \bf{IT} & \bf{Koran} & \bf{Law} & \bf{Medical} & \bf{Subtitles} \\ \hline
\rm Train & $223$K & $18$K& $467$K& $248$K& $14.46$M\\
\rm Valid & $2$K & $2$K& $2$K& $2$K& $2$K\\
\rm Test & $2$K & $2$K& $2$K& $2$K& $2$K\\
\hdashline
\rm Size & $3.6$M & $0.5$M& $19.1$M& $6.9$M& $180.7$M\\
\hline
\end{tabular}}
\caption{
    The statistics of datasets in different domains. We also list the size of the datastore, which is the number of stored pairs.
	}\label{tab_stat}
\end{table}

\section{Effect of Datastore Size}
\begin{figure}[!t]
	\begin{center}
	\includegraphics[width=1.0\linewidth]{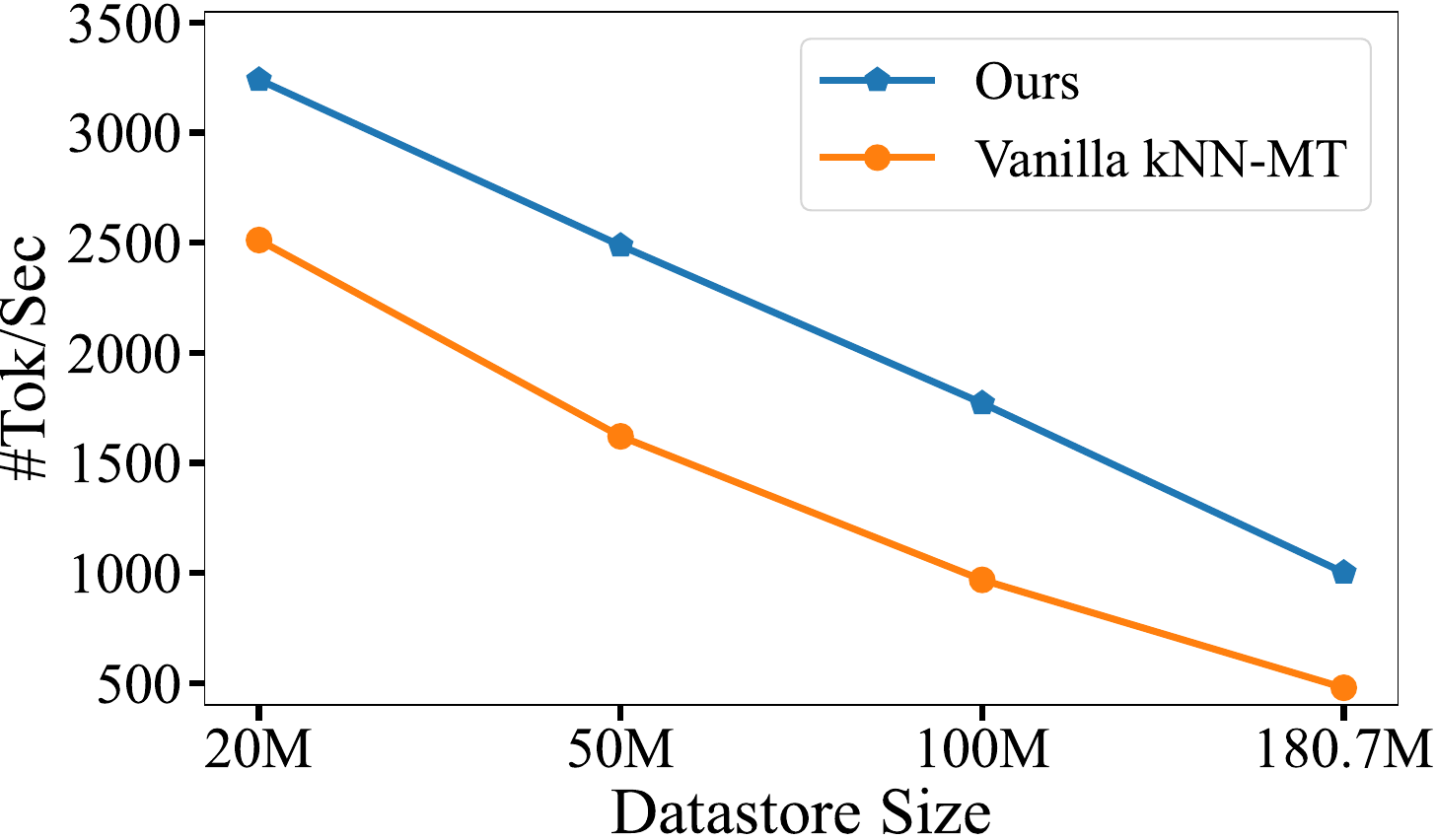}
	\end{center}
 	\caption{ Decoding speed(\#Tok/Sec$\uparrow$) of Vanilla $k$NNMT and ours. Here, we set the batch size as $128$.
	}\label{tab_datastoresize_decoding_speed}
\end{figure}

As analyzed in Section~\ref{analyses_of_decoding_speeds}, our speed advantage becomes more significant with the increase of datastore size. To further verify this, we construct datastores of varying sizes by randomly deleting pairs from the original datastore, and employ the pruned datastores for $k$NN retrieval. The results of decoding speed on the Subtitles dataset are reported in Figure~\ref{tab_datastoresize_decoding_speed}.
As expected, we observe that our model consistently surpasses $k$NN-MT, regardless of the datastore size. Furthermore, the efficiency advantage of our model over $k$NN-MT becomes more evident with the increase of datastore size. These results further confirm that the pronounced speed advantage of our model as the datastore expands.

\section{Hyper-Parameter Tuning}
\label{hyper-parameter_tuning}
The performance and efficiency of our model is significantly impacted by the hyper-parameter $\alpha_{min}$, and we tune $\alpha_{min}$ among the subset of $\{0.45, 0.40, 0.35\}$ on the validation set. 

We report the BLEU scores and \#Tok/Sec, as shown in Table~\ref{tab_hyper_parameter}. As $\alpha_{min}$ decreases, the increment in BLEU scores gradually diminishes, while the drop in decoding speed becomes more pronounced. So we set the hyper-parameter $\alpha_{min}$ to $0.45$ for Koran, Subtitles, and $0.40$ for other datasets to achieve a balance between performance and efficiency.

Note that as the validation set is utilized in training the classifier network, there exists a potential risk of overfitting when tuning $\alpha_{min}$, which may result in a suboptimal selection of $\alpha_{min}$.

\begin{table}[!h]
	\renewcommand
	\arraystretch{1.15}
	\centering
\resizebox{1.0\linewidth}{!}{
\begin{tabular}{l|ccccc}
\hline
\bf{Datasets} & \bf{0.45} & \bf{0.40} & \bf{0.35} \\ \hline
IT & $42.03$ / $2978.73$ & $42.30$ / $2940.88$ & $42.23$ / $2878.88$\\
Koran & $19.53$ / $3452.23$ & $19.50$ / $3415.35$ & $19.58$ / $3408.09$\\
Law  & $58.66$ / $3137.26$ & $59.20$ / $3097.68$ & $59.31$ / $3001.18$\\
Medical & $51.45$ / $3155.22$ & $51.75$ / $3069.02$ & $51.86$ / $2989.31$\\
Subtitles & $32.05$ / $1027.21$ & $32.13$ / $898.61$ & $32.09$ / $771.34$\\
\hline
\end{tabular}
}
\caption{
  BLEU$\uparrow$ and \#Tok/Sec$\uparrow$ of our model on the multi-domain validation sets with different $\alpha_{min}$. Here, we set the batch size as $128$.
	}\label{tab_hyper_parameter}
\end{table}

\begin{table*}[!h]
	\renewcommand
	\arraystretch{1.2}
	\centering
\resizebox{1.0\linewidth}{!}{
\small
\begin{tabular}{l|ccccc|c}
\hline
\bf{Model} & \bf{IT} & \bf{Koran} & \bf{Law} & \bf{Medical}\\ 
\hline
INK & $49.06$ / $2842.24$ & $22.35$ / $3401.44$ & $63.51$ / $2922.82$ & $57.41$ / $2687.21$\\
\hline
INK with Robust $k$NN-MT & $49.97$ / $1489.68$ & $20.90$ / $1839.12$ & $65.41$ / $1053.55$ & $58.30$ / $1391.53$ \\
\rowcolor{lightgray}
\quad + Ours & $49.72$ / $2065.52$ & $21.40$ / $2243.05$ & $65.17$ / $1734.07$ & $57.98$ / $1788.20$\\
\hline
\end{tabular}
}
\caption{
    BLEU$\uparrow$ and \#Tok/Sec$\uparrow$ of models on the multi-domain test sets. We are unable to provide the results on the Subtitles domain, since INK needs to fine-tune the base NMT model and reconstructs the datastore at each epoch, which is extremely time-consuming on the Subtitles domain.
	}\label{compatibility_ink}
\end{table*}

\section{Compatibility with INK}
\label{compatibility_with_ink}
INK~\cite{zhu2023ink} achieves excellent performance by performing parameter-efficient fine-tuning on the base NMT model using domain-specific data through knowledge distillation, and its variant equipped with Robust $k$NNMT~\cite{jiang-etal-2022-towards} achieves the state-of-the-art performance, we conduct compatibility experiments on this variant with our model and report the BLEU scores and decoding speed as shown in Table~\ref{compatibility_ink}. We can observe that our model can improve the efficiency with only a slight drop in translation quality. Thus, we confirm that our model is compatible with INK.

\end{document}